\documentclass[sigconf]{acmart}
\AtBeginDocument{%
  \providecommand\BibTeX{{%
    \normalfont B\kern-0.5em{\scshape i\kern-0.25em b}\kern-0.8em\TeX}}}

\setcopyright{acmlicensed}
\copyrightyear{2018}
\acmYear{2018}
\acmDOI{XXXXXXX.XXXXXXX}

\acmConference[Conference acronym 'XX]{Make sure to enter the correct
  conference title from your rights confirmation emai}{June 03--05,
  2018}{Woodstock, NY}
\acmISBN{978-1-4503-XXXX-X/18/06}

\newcommand{\vx}{\mathbf{x}}

\newcommand{\vthe}{\boldsymbol{\theta}}

\newcommand{\mH}{\mathcal{H}}

\newcommand{\vY}{\boldsymbol{Y}}

\newcommand{\vgamma}{\boldsymbol{\gamma}}

\newcommand{\E}{\mathbb{E}}

\usepackage{soul}
\usepackage{xcolor}

\newcommand{\Mean}{{\mathbb{E}}}

\newcommand{\Cov}{\boldsymbol{\Sigma}}

\newcommand{\prob}{\mathbb{P}}

\newcommand{\vPsi}{\boldsymbol{\Psi}}

\newcommand{\vZ}{\boldsymbol{Z}}

\newcommand{\vmu}{\boldsymbol{\mu}}

\usepackage{amsthm}
\theoremstyle{plain}

\newlength{\leftstackrelawd}
\newlength{\leftstackrelbwd}
\def\eqstack#1#2{\settowidth{\leftstackrelawd}%
{${{}^{#1}}$}\settowidth{\leftstackrelbwd}{$#2$}%
\addtolength{\leftstackrelawd}{-\leftstackrelbwd}%
\leavevmode\ifthenelse{\lengthtest{\leftstackrelawd>0pt}}%
{\kern-.5\leftstackrelawd}{}\mathrel{\mathop{#2}\limits^{#1}}}

\usepackage[utf8]{inputenc}
\usepackage{array}
\usepackage{amsmath}
\usepackage{bm}
\usepackage{graphicx}
\usepackage{booktabs}
\usepackage{setspace}
\usepackage{textcomp}
\usepackage{enumitem}
\usepackage{datetime2}
\usepackage{multirow}
\usepackage[font=small,labelfont=bf]{caption}
\usepackage{todonotes}

\usepackage{subcaption}

\usepackage{xcolor}
\usepackage{dsfont}

\usepackage{lineno}
\usepackage[ruled,vlined]{algorithm2e}
\setcitestyle{square,comma}

\begin{document}

%%
%% The "title" command has an optional parameter,
%% allowing the author to define a "short title" to be used in page headers.
\title{Multi-Task Combinatorial Bandits for Budget Allocation}

%%
%% The "author" command and its associated commands are used to define
%% the authors and their affiliations.
%% Of note is the shared affiliation of the first two authors, and the
%% "authornote" and "authornotemark" commands
%% used to denote shared contribution to the research.
\author{Lin Ge}
\affiliation{%
  \institution{North Carolina State University}
    \city{Raleigh}
  \state{NC}
  \country{USA}
}
\author{Yang Xu}
\affiliation{%
  \institution{North Carolina State University}
    \city{Raleigh}
    \state{NC}
  \country{USA}
}
\author{Jianing Chu}
\affiliation{%
  \institution{North Carolina State University}
    \city{Raleigh}
    \state{NC}
  \country{USA}
}

\author{David Cramer}
\affiliation{%
  \institution{Amazon}
    \city{Seattle}
    \state{WA}
  \country{USA}
}

\author{Fuhong Li}
\affiliation{%
  \institution{Amazon}
    \city{Seattle}
      \state{WA}
  \country{USA}
}

\author{Kelly Paulson}
\affiliation{%
  \institution{Amazon}
    \city{Seattle}
      \state{WA}
  \country{USA}
}

\author{Rui Song}
\affiliation{%
  \institution{Amazon}
  \city{Seattle}
      \state{WA}
  \country{USA}
}

%%
%% By default, the full list of authors will be used in the page
%% headers. Often, this list is too long, and will overlap
%% other information printed in the page headers. This command allows
%% the author to define a more concise list
%% of authors' names for this purpose.
\renewcommand{\shortauthors}{Ge, et al.}

%%
%% The abstract is a short summary of the work to be presented in the
%% article.
\begin{abstract}
Today's top advertisers typically manage hundreds of campaigns simultaneously and consistently launch new ones throughout the year. A crucial challenge for marketing managers is determining the optimal allocation of limited budgets across various ad lines in each campaign to maximize cumulative returns, especially given the huge uncertainty in return outcomes. In this paper, we propose to formulate budget allocation as a multi-task combinatorial bandit problem and introduce a novel online budget allocation system. The proposed system: i) integrates a Bayesian hierarchical model to intelligently utilize the metadata of campaigns and ad lines and budget size, ensuring efficient information sharing; ii) provides the flexibility to incorporate diverse modeling techniques such as Linear Regression, Gaussian Processes, and Neural Networks, catering to diverse environmental complexities; and iii) employs the Thompson sampling (TS) technique to strike a balance between exploration and exploitation. Through offline evaluation and online experiments, our system demonstrates robustness and adaptability, effectively maximizing the overall cumulative returns. A Python implementation of the proposed procedure is available at https://anonymous.4open.science/r/MCMAB.
\end{abstract}


\received{20 February 2007}
\received[revised]{12 March 2009}
\received[accepted]{5 June 2009}

\maketitle

\section{Introduction}
Budget allocation has been given wide attention in the advertising market \cite{chiu2018optimal,luzon2022dynamic}. Advertisers and agencies that use Demand Side Platforms (DSP), like Amazon DSP, routinely manage hundreds of simultaneous campaigns, each comprising various ad lines targeting specific audiences and set with diverse delivery settings. Daily, marketing managers allocate budgets across ad lines for each campaign within a daily budget, aiming to boost traffic to retail websites or maximize product sales. A key challenge is the lack of understanding of the relationship between ads spending and performance outcomes, which, once obtained, reduces the task to an optimization problem solvable by various methods \cite{karlsson2020feedback, geng2021automated}.

%It is always crucial for decision makers to learn an optimal strategy that can optimize the overall reward given limited resources such as cash flow, time and labor.

Today's ADSP implements automated online performance optimizations that respond to signals on each campaign's own spend and conversion data. Such an approach often encounters two significant challenges: \textbf{First}, advertisers frequently launch campaigns sequentially or run multiple campaigns simultaneously. Without a design in the ADSP that effectively coordinates learning from past and concurrent campaigns, the learning process is inefficient. Ad lines typically begin with an equal budget distribution at the start of a campaign. Although automated tools can make real-time budget adjustments during campaigns, determining the optimal budget allocation mix can take up to three weeks. This delay would result in a considerable number of ineffective ad deliveries, a general issue particularly severe for short-lived campaigns and large advertisers or agencies participating in numerous campaigns each year. \textbf{Second}, measurement uncertainty exists, primarily due to the lack of real-life counterfactual observations, necessitating the use of estimations. This challenge is exacerbated by the dynamic nature of advertising and business environments, where factors like seasonality, competitive bidding, privacy regulations, DSP functionality, and fluctuating customer behaviors make it even more complex to accurately estimate these values. Relying solely on historical data can lead to suboptimal solutions in sequential decision-making scenarios.
Thus, the following question is addressed in this paper: \textbf{\textit{How can we wisely utilize i) learnings from past campaigns and ii) insights from other ongoing concurrent campaigns to accelerate the learning process for optimal budget allocation?}} 

\begin{figure}
    \centering
    \includegraphics[width=\linewidth]{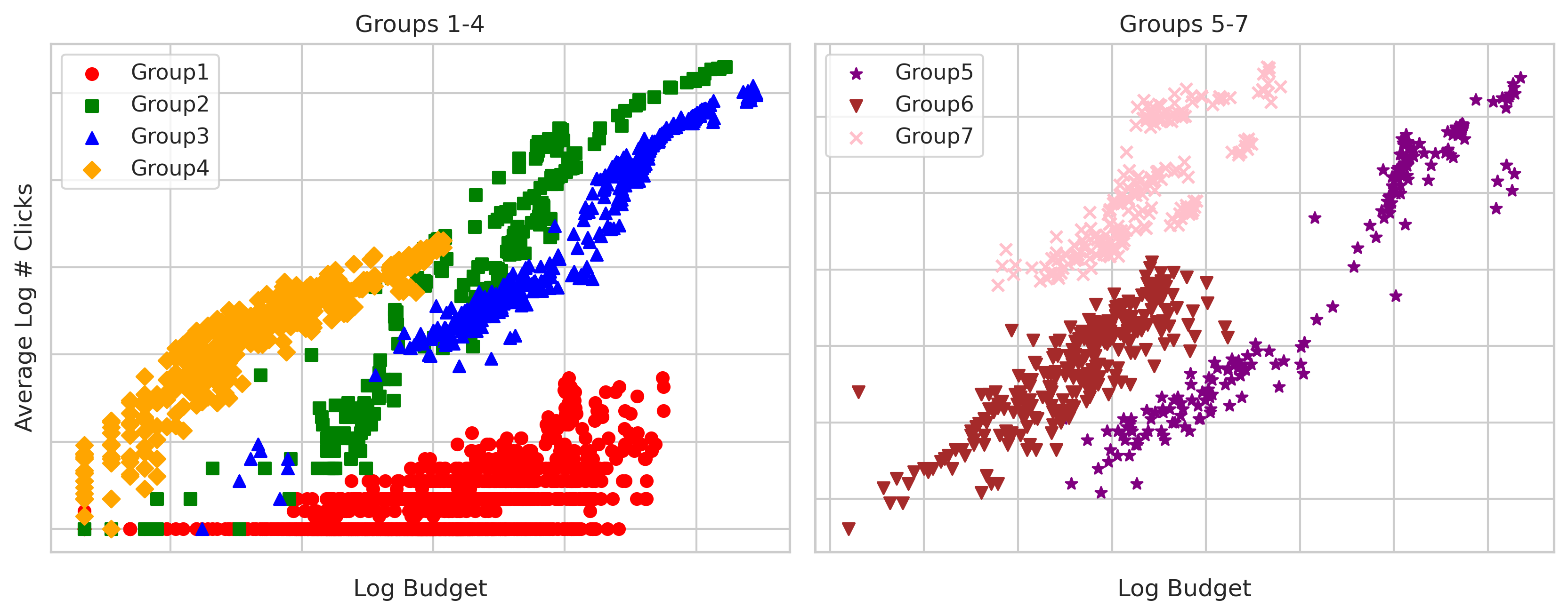}
    \vspace{-.4cm}
    \caption{Scatter plots of the log of average number of clicks received and the log of budget allocated for various \textit{\textbf{ad line groups}} with distinct advertisers' industries, channels, supply sources, and audiences.}
    \label{fig:motivation}
    \vspace{-.3cm}
\end{figure}

To address the inherent uncertainty in digital advertising, Bandits algorithms are known for effectively balancing exploration and exploitation and have recently been applied to budget allocation, formalizing it as a combinatorial multi-armed bandit (CMAB) problem \citep{zuo2021combinatorial, nuara2018combinatorial, nuara2022online}. However, none of them investigated how to share information across multiple ad lines and campaigns. Figure \ref{fig:motivation} depicts the average clicks received for different ad lines from various campaigns, with varying levels of budget assigned, underscoring the significant influence of ad/campaign characteristics on budget-performance relationships. Additionally, according to Figure \ref{fig:motivation}, it is important to note that the available features do not fully determine the expected performance of an ad line with an assigned specific budget. In other words, even when conditioned on informative features, the expected performance for a given budget level still exhibits a degree of variability (referred to as \textit{inter-arm heterogeneity}). To strategically boost information sharing across tasks, we model budget allocation in ADSP as a multi-task CMAB problem where each ad campaign represents a single CMAB task, and introduce a feature-based Bandit algorithm augmented by a Bayesian hierarchical model.% designed to boost information sharing across both ad campaigns and ad lines effectively.

\textbf{Contributions.} Our main contributions are multi-fold. 

\textbf{First}, our study is the first, to our knowledge, to investigate budget allocation for online advertising from the perspective of large advertisers and agencies managing multiple campaigns, with a meta objective of optimizing performance over the campaign distribution. This contrasts with existing studies that primarily focus on optimizing a single campaign from an advertiser's viewpoint \citep{han2021budget,nuara2018combinatorial, nuara2022online}.

\textbf{Second}, to capture the budget-performance relationship, we propose a general Bayesian hierarchical model that supports both parametric and non-parametric modeling. Driven by similar motivations, \citet{han2021budget} introduced a contextual bandit-based system using augmented data generated from a global model to share information across ad lines. However, not accounting for the uncertainty of the fitted global model, its effectiveness relies heavily on the global model's performance and can lead to suboptimal decisions, especially in data-limited scenarios. Furthermore, with the power law assumption, they assume a linear relationship between the logarithms of the budget and the performance metric, which often fails in practice, as shown in Figure \ref{fig:motivation}. In contrast, our method can effectively address more complex nonlinear budget-performance relationships by integrating with the Gaussian Process and Neural Network, as two instances.

\textbf{Third}, through the construction of the Bayesian hierarchical model, which incorporates an arm-specific random effect to capture the information not being explained by the feature information, we effectively tackle the inter-arm heterogeneity observed in Figure \ref{fig:motivation}. While none of the aforementioned work addresses this ubiquitous issue, recent advancements in meta bandits \citep{wan2021metadata, wan2023towards} likewise focus on dealing with such heterogeneity. However, they all rely on parametric linear model assumptions. %Our approach distinguishes itself by allowing for non-parametric modeling, thereby relaxing these conventional assumptions.

\textbf{Finally}, we implemented our proposed framework in an offline study using real campaign data from ADSP. The results consistently indicate that our framework achieves faster convergence and higher cumulative reward, thereby leading to a better budget allocation strategy in the long term. This is also supported by an online experiment. %An online experiment is planned for 2024.

\section{Related Work}\label{related work}
Budget/resource allocation, extensively explored over the past decades, has recently been formalized within the framework of CMAB. %bandits problems. Depending on how budget constraints are set, the relevant literature can be categorized into two main groups: 1) CMAB, and 2) bandits with knapsack.
Formulating the allocation problem as CMAB \cite{xu2019efficient, xu2020combinatorial, zuo2021combinatorial, gupta2022correlated, nuara2018combinatorial, nuara2022online}, budgets are discretized into finite proportions, aligning with the nature of combinatorial bandits in slate recommendation. Notably, \citet{zuo2021combinatorial} leveraged the CMAB framework by defining a super arm as an (ad line, budget) tuple for action assignment, and naturally extended this idea to continuous budget allocation scenarios with additional Lipschitz continuity assumptions. In the work by \citet{xu2020combinatorial}, the authors studied resource allocation problem
with concave objective and fairness constraints. \citet{nuara2018combinatorial, nuara2022online} proposed a joint bid/budget optimization algorithm based on Bayesian bandits update with Gaussian process. In the work by \citet{gupta2022correlated}, the authors proposed a correlated combinatorial bandit framework to capture the structural correlations between reward functions. %Although \citet{nuara2018combinatorial, nuara2022online} addressed the budget allocation problem in multiple campaigns, each subcampaign used an independent bandit update procedure, revealing a lack of information sharing in the online learning process. 
However, these methods either fail to utilize contextual information \citep{xu2019efficient,zuo2021combinatorial} or rely on restrictive modeling assumptions in rewards and resource consumptions \citep{xu2020combinatorial}, limiting their applicability to more general applications.

%In order to share information across campaigns and ad lines to foster bandits learning, 
To utilize the contextual information expediting the learning process, \citet{han2020contextual} introduced a contextual bandit framework via a global-local model. However, \citet{han2020contextual} focuses on a single budget allocation task and ignores the inter-arm heterogeneity. Additionally, their updating procedure lacks effective exploration of the global model, making the methodology's performance heavily dependent on how well the global model fits the data. This poses a potential challenge when dealing with limited sample sizes or significant noise in the data. In this work, we build upon the combinatorial bandit framework to do information sharing. While discretizing budgets in CMAB may introduce a minor bias in the precision of estimating the optimal arm, this approach eliminates the necessity of imposing smoothness assumptions, which is typically required when the budget is considered continuous \cite{han2020contextual}. %To the best of our knowledge, the majority of existing research, despite incorporating multiple subcampaigns/adlines into a unified budget allocation problem, has traditionally approached this issue from a single-task perspective. This stands in contrast to our scenario, where our objective is to concurrently allocate budgets across ad lines for multiple campaigns. While this context is prevalent in demand-side platforms, there is currently no established method within a meta-bandits framework to address this unique requirement. 
Additionally, our work is closely related to traditional approaches that consider budget allocation purely as an optimization problem without addressing estimation uncertainty \citet{ou2023survey}, and the body of work on Multi-task/Meta Bandits \citep{wan2021metadata}. %For additional related works, refer to Appendix \ref{additional related work}.

\section{Preliminaries}
%\vspace{-.1cm}
\subsection{Budget Allocation}
Consider a large advertiser hosting a collection of $M$ advertising campaigns running either concurrently or sequentially.  In each campaign $m\in[M]=\{1,2,…,M\}$, there are $K_m$ ad lines designated for daily budget allocation, and the campaign duration is denoted as $T_m$. For a given campaign $m$, we assume a total daily budget of $B_m$. At each time round $t$, the budget assigned to each ad line $k\in[K_m]$ is denoted as $a_{m,k,t}$ and  a constraint exists such that the sum of the assigned budgets across all ad lines in campaign $m$ satisfies $\sum_{k=1}^{K_m}a_{m,k,t}\leq B_m$. After assigning budget $a_{m,k,t}$ for each ad line $k$ in campaign $m$, we consequently observe a random reward $R_{m,k,t}(a_{m,k,t})$. Our goal is to maximize the cumulative reward function across all ad lines, all campaigns and all rounds:
\begin{equation}\label{eq:opt}
    \begin{aligned} 
   & \mathop{\rm{maximize}}\limits_{a_{m,k,t}} \sum_{m=1}^M\sum_{t=1}^{T_m} \sum_{k=1}^{K_m}\mathbb{E}\{R_{m,k,t}(a_{m,k,t})\},\\ 
    &\text{subject to} \sum_{k=1}^{K_m}a_{m,k,t} \leq B_{m}   ~\ \forall m, t.
    \end{aligned}
\end{equation}

%\vspace{-.2cm}
\subsection{Combinatorial Multi-Armed Bandits}
To connect the optimization problem described above with the framework of CMAB, we begin by introducing the fundamental concept of CMAB. A typical CMAB problem consists of $K$ arms associated with a set of random variables $R_{k,t}$.  The random variable $R_{k,t}$ indicates the random reward of arm $k\in[K]$ at time round $t$. The set of all possible subsets of arms is a power set $\mathcal{S}=2^{[K]}$. We refer to every set of arms $S\in \mathcal{S}$ as a super arm and every arm in $S$ as a base arm. At each time round, one super arm $S\in\mathcal{S}$ is played and the rewards of all base arms in this super arm are observed. 

We consider each campaign as a single-task CMAB problem where each ad line within the campaign corresponds to a base arm. The first challenge is that, unlike the conventional CMAB problem where the decision revolves around whether to play the base arm or not, in our scenario, we must also determine the budget allocation for each chosen base arm. As the budget amount is continuous, we confront an infinite number of potential base arms. The second challenge arises from the presence of daily budget constraints. This implies that, at each round, only a subset of super arms is viable for play, restricted by the limitations imposed by the available daily budget. Even upon overcoming these challenges, conventional CMAB are designed to accommodate only a single campaign. However, advertisers often initiate new campaigns or run multiple campaigns concurrently. To mitigate the cold start issue associated with new campaigns and enhance data utilization, the proposed CMAB must facilitate information sharing across different campaigns.

%\vspace{-.2cm}
\section{Methodology}
%To tackle the challenges outlined above using CMAB, we propose the following multi-task Bayesian hierarchical CMAB framework.
%\vspace{-.1cm}
\subsection{Problem Formulation}
Without loss of generality, we define the continuous action space as $\mathcal{A}=[0,1]$, where $a_{m,k,t}\in \mathcal{A}$ represents the proportion of the total budget allocated to ad line $k$ in campaign $m$ at time round $t$. The corresponding budget can be expressed as $B_{m} \times a_{m,k,t}$.  We further discretize the action space by partitioning the continuous budget into different proportions: $\mathcal{A}_d=\{0,\frac{1}{N},\frac{2}{N},\cdots,\frac{N-1}{N},1\}$, with $N$ denoting a user-specified integer constant. The rationale behind the discretization is twofold. First, in practical scenarios, campaign budgets are commonly assigned in rounded percentages, such as 10\% or 25\%, rather than extremely precise amounts. Second, discretization eliminates the need for smoothness assumptions typically required for continuous budget optimization. %Second, this discretization results in a finite set of base arms, enabling us to analyze the problem under the CMAB framework. 
Accordingly, we consider maintaining a set of base arms in campaign $m$ as $\{(k,a)\mid k\in[K_m],a\in \mathcal{A}_d\}$. 
Each base arm contains metadata $x_{m,k}$,which comprises specific information on campaign and ad line configurations. At each time round $t$, we can only play one arm $(k,a)$ for each ad line $k$ in campaign $m$. This implies the allocation of a budget proportion $a_{m,k,t}$ to ad line $k$ in campaign $m$. Let $\theta_{m,k,a}\equiv \E\{R_{m,k}(a)\}$. We can rewrite our goal as:
    \begin{align}\label{eq:reform}
 \mathop{\rm{maximize}}\limits_{a_{m,k,t}} &  \sum_{m=1}^M\sum_{t=1}^{T_m} \sum_{k=1}^{K_m}\theta_{m,k,a_{m,k,t}},\\ 
    \text{subject to}& \sum_{k=1}^{K_m} a_{m,k,t}\leq 1,    \; \sum_{n=0}^N I\left(a_{m,k,t}=\frac{n}{N}\right)=1, ~\ \forall m, t.\nonumber
    \end{align} Thus, let $\boldsymbol{a}_{m,t} = (a_{m,1,t}, \cdots, a_{m,K_m,t})$, the full allocation space for each campaign $m$ at each decision point is $\mathcal{S}_m = \{\boldsymbol{a}_{m,\cdot} \mid a_{m,k,\cdot}\in\mathcal{A}_d, \sum_{k=1}^{K_m} a_{m,k,\cdot}\leq 1, \sum_{n=0}^N I\left(a_{m,k,\cdot}=\frac{n}{N}\right)=1, \forall k \}$.
%  We consider the following model:

% \begin{align*}
%             Y_{m,k,t}  &=  f(Y_{m,k,t}\mid x_{m,k},a_{m,k,t},\theta_{m,k,a}), \\
%             R_{m,t} &= \sum_{k=1}^{K_m}Y_{m,k,a,t},
% \end{align*}

% where $Y_{m,k,a,t}$ is the observed reward for each base arm and $R_{m,t}$ is the observed reward for the super arm $S_{m,t}$. The observation $Y_{m,k,a,t}$ is generated following a domain model $f$, which is parameterized by some unknown parameters $\theta_{m,k,a}$. In real problems, $f$ is typically a complex distribution and may involve nonlinear
% functions, $\theta_{m,k,a}$ can be high-dimensional, and the sets of super arms can be huge.
%\vspace{-.1cm}
\subsection{Multi-task Bayesian Hierarchical CMAB Framework}
%To solve the optimization problem (\ref{eq:reform}), an important step is to estimate $\theta_{m,k,a}$. To share information across campaigns and ad lines while accounting for inter-arm heterogeneity, we consider the following Bayesian hierarchical model, 
        \begin{figure}
    \vspace{-.3cm}
        \centering
        \includegraphics[width = \linewidth]{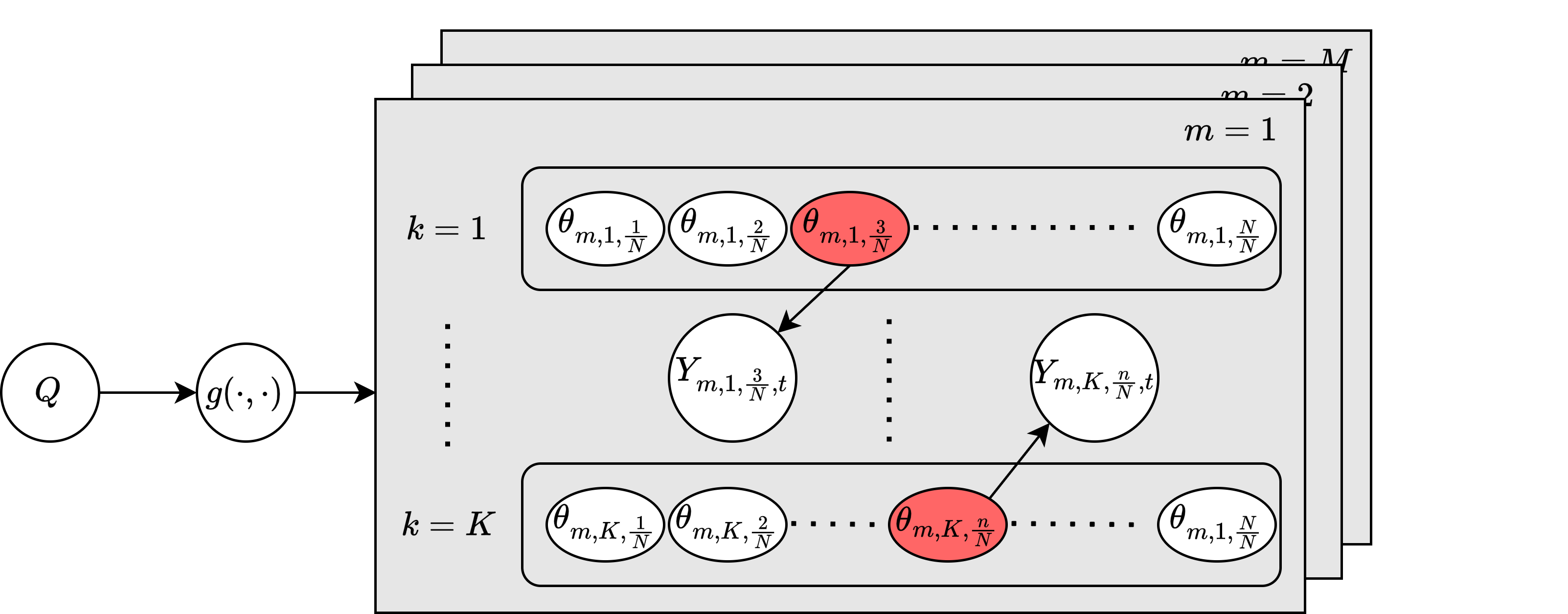}
        \vspace{-.3cm}
        \caption{Graphical representation of model (\ref{eqn:hierachical_model}). Red nodes are the selected base arm at round $t$. }
   \label{fig:hiearchical_model}%$Q$ denotes the prior information related to $g$.}
   \vspace{-.3cm}
\end{figure}
To tackle the optimization problem (\ref{eq:reform}), a key step is the estimation of $\theta_{m,k,a}$, which we propose to accomplish using the following Bayesian hierarchical model:
\begin{equation}\label{eqn:hierachical_model}
            \begin{alignedat}{2}
            &\text{(Prior)} \quad \quad \quad \quad \quad
                \;    \text{Prior information Q} && \text{ related to} ~\ g,\\
                &\text{(Generalization)} \quad
                \;    \theta_{m,k,a}\mid \vx_{m,k},a&&= g(\vx_{m,k},a) +\delta_{m,k,a},\\
                & &&\quad
                \forall m\in[M],k\in[K_m],a\in\mathcal{A}_d, \\
                &\text{(Observation)} \quad
                \;\quad\quad \;   Y_{m,k,a_{m,k,t},t}  &&=  \theta_{m,k,a_{m,k,t}} + \epsilon_{m,k,t}, \\
                %& &&\quad \quad\quad \forall  m\in[M], k\in[K_m],\\
                &\text{(Reward)} \quad \quad\quad
                \quad\quad \quad \; R_{m,t} &&= \sum_{k\in[K_m]}Y_{m,k,a_{m,k,t},t}, %\sum_{(k,a)\in A_{m,t}}Y_{m,k,a,t}.
            \end{alignedat}
        \end{equation}
        where $\theta_{m,k,a}$ is the expected reward of allocating a budget of $B_m * a$ to ad line $k$ in campaign $m$ and $\epsilon_{m,k,t} \sim N(0, \sigma_\epsilon^2)$ is the random noise for some known $\sigma_\epsilon$. At round $t$, $Y_{m,k,a,t}$ is the observed reward for ad line $k$ and $R_{m,t}$ is the total reward aggregating the observed rewards from all ad lines within campaign $m$. The essence of (\ref{eqn:hierachical_model}) lies in the two-way decomposition of $\theta_{m,k,a}$, which splits $\theta_{m,k,a}$ into two components: i) $g(\vx_{m,k}, a)$, a function capturing the average impact of available features $\vx_{m,k}$ and action $a$ on the reward, with $Q$ as the prior belief about $g$'s distribution, and ii) $\delta_{m,k,a}$, a random effect that accounts for the inter-arm heterogeneity conditioned on $\vx_{m,k}$ and $a$. See Figure \ref{fig:hiearchical_model} for an illustration. Recognizing that even the most advanced machine learning algorithms cannot perfectly represent the relationship between features and reward, %leading to inevitable model misspecification, 
        the additional random effect is primarily employed to account for the uncertainty in $\theta_{m,k,a}$ that failed to be captured by $g$. Intuitively, as such, we utilize i) the shared information across campaigns and ad lines via $g$ and ii) the observations $Y_{m,k,a_{m,k,t},t}$ from all correlated base arms, to infer $\theta_{m,k,a}$.

        Let $\boldsymbol{\delta}_m = \big[\delta_{m,1, 0}, \cdots, \delta_{m,K,\frac{N}{N}}\big]$, we assume that $\boldsymbol{\delta}_m \sim N(\boldsymbol{0}, \Cov)$ for some known covariance matrix $\Cov$. 
%The parameter $\theta_{m,k,a}$ is derived from a combination of a function $g$ and a random effect $\delta_{m,k,a}$. The observation $Y_{m,k,a_{m,k,t},t}$ is generated by $\theta_{m,k,a_{m,k,a}}$  and a random noise term $\epsilon_{m,k,t}\sim N(0, \sigma_\epsilon^2)$ for some known $\sigma_\epsilon$. Specifically, $g$ is an unknown function of  $\vx_{m,k}$ and $a$ 
%capturing the average influence  of available features and action on the reward, and $\delta_{m,k,a}$ is the base-arm-specific random effect modeling the inter-arm heterogeneity. 
 For $g$, either a parametric or non-parametric model can be used. In this work, we consider three working models as examples: i) Linear regression (LR), assuming $g(\vx_{m,k}, a) = \phi(\vx_{m,k}, a)^T\vgamma$, where $\phi(\vx_{m,k},a)$ is a certain transformation of features and actions, and $\vgamma$ is a vector of parameters with a prior $\vgamma\sim N(\vmu_{\vgamma}, \Cov_{\vgamma})$. ii) Neural network (NN) regression, considering $g(\vx_{m,k},a)$ as a fully connected NN of depth $L\geq 2$, the collection of parameters of which, $\vgamma$, has a prior $\vgamma\sim N(\vmu_{\vgamma}, \Cov_{\vgamma})$ \citep{jacot2018neural}. iii) Gaussian process (GP) regression, assuming that $g(\vx_{m,k},a)$ follows a Gaussian process prior, such that $g \sim \text{GP}(\mu_{\vgamma},\mathcal{K}_{\vgamma})$.
\subsection{Learning Strategy}
\subsubsection{Posterior Distributions}

%\textcolor{red}{we adopt the meta-learning viewpoint \citep{vilalta2002perspective} (i.e., learning how to learn efficiently for each base arm).}

% 

To sequentially update the parameter estimation in an online setting, a key step is to derive the posterior distribution of parameters in (\ref{eqn:hierachical_model}). Let $\vthe_m = \{\theta_{m,k,a}\}_{1\leq k\leq K_m, 1\leq a\leq N}$ and $\vthe = \{\vthe_m\}_{1\leq m\leq M}$ as a $\sum_{m\in[M]}K_mN$-dimensional vector containing the expected reward for all $(m,k,a)$ tuples. 
%We proposed a two-step iteration procedure via Thompson sampling to update the posterior of $\vthe$ given the constantly updating historical data $\mH$.  
Since
$\mathds{P}(\vthe\mid \mH)\propto \mathds{P}(\vthe\mid \mH,g)\mathds{P}(g\mid \mH)$,
we split the posterior derivation into two parts: 1) $\mathds{P}(g\mid\mH)$, and 2) $\mathds{P}(\vthe\mid\mH,g)$. 

$\mathds{P}(g\mid\mH)$ shares a general structure under LR, GP and NN. Let $\vPsi_{1:H}$ be a $d \times KH$ matrix comprising features of the $K$ selected arms (one for each ad line) offered from round $1$ to round $H$. Similarly, $\vY_{1:H} = (\vY_{1}^{T},\cdots,\vY_{H}^{T})^{T}$ denotes the observed rewards of all base arms offered up to round $H$. $\mathds{P}(g\mid\mH)$ follows a normal distribution, with mean and covariance as
\begin{align*}
    & \mathbb{E}(g(\cdot)\mid\mH) =\mu(\cdot) + \mathcal{K}(\cdot,\vPsi_{1:H})(\Phi+\sigma^2 I )^{-1} (\vY_{1:H}-\mu(\vPsi_{1:H}))\\
    & \text{Cov}(g(\cdot)\mid\mH) = \mathcal{K}(\cdot,\cdot) - \mathcal{K}(\cdot,\vPsi_{1:H})(\Phi+\sigma^2 I )^{-1} \mathcal{K}(\vPsi_{1:H},\cdot),
\end{align*}
where $\Phi=\mathcal{K}(\vPsi_{1:H},\vPsi_{1:H})+ \Sigma_{1:H}$. Here, $\mathcal{K}(\vPsi_{1:H},\vPsi_{1:H})$ is the variance induced by the prior distribution, and $\Sigma_{1:H}$ denotes the variance induced by the random effect. In LR, $\mu(x) = x^T\vgamma$ takes a specific linear form, and $\mathcal{K}(x,x') = x^T\Sigma_{\vgamma} x'$. In GP, $\mu(x)$ as the prior mean can adopt any function form of $x$, and $\mathcal{K}(x,x')$ is a general kernel function. It can be a linear kernel $\mathcal{K}(x,x')=\langle x,x'\rangle$, an RBF kernel $\mathcal{K}(x,x')= \exp{-|x-x'|^2/(2l^2)}$ with $l$ as a hyperparameter representing the standard deviation, or other kernel functions. In NN, $\mu(x)$ is a fully-connected neural network, and $\mathcal{K}(x,x')$ is the neural tangent kernel.

Given $g$, $\mathds{P}(\vthe_m\mid\mH,g)$ follows a normal distribution with
% \textit{Step 2}: We sample $\widehat{R}_{m,k,t}(a_{m,k,t})$ from the posterior distribution of $\theta_{m,k,a}$, 
% where 
\begin{align*}
    &\mathbb{E}(\vthe_m\mid\mH, g) =\text{Cov}(\vthe_m\mid\mH, g)
    \left[\Cov^{-1}g(\vPsi_m) + \sigma^{-2}\vZ_{1:H,m}\vY_{1:H}\right],\\
    &\text{Cov}(\vthe_m\mid\mH, g) = \left(\Cov^{-1}+\sigma^{-2}diag(C_{m,1,1},\cdots, C_{m,K,N})\right)^{-1},
\end{align*} where $C_{m,k,n}$ is the number of observations in $\mH$ that correspond to base arm $(m,k,n)$.

%In practice, instead of updating both posteriors for every individual data point, we employ a batch updating process to expedite the posterior updating. Specifically, given the high computational complexity, the posterior of $g\mid\mH$ is updated only at the start of each round. The posterior of $\vthe\mid\mH,g$ is updated whenever a new sample is observed, following the same procedure as before. Further algorithm acceleration details are provided in Appendix \ref{appendix:batch_memory}.

\subsubsection{TS and Optimization}
Using the derived posteriors, we adopt the classical TS-type algorithm but split the posterior sampling into two steps for each decision point. Specifically, we first sample $\tilde{g} \sim \mathds{P}(g\mid \mH)$ and then sample $\tilde{\vthe}_m \sim \prob(\vthe_m\mid\tilde{g}, \mH)$. In the first step, we integrate all collected data to create a feature-based informative prior for each $\theta_{m,k,a}$, which then guides the subsequent learning of $\theta_{m,k,a}$. We refer to this as a feature-guided (FG) approach. This notably differs from that of \cite{han2021budget}, which employs only the first step. 

%Such a two-step sampling approach enables us to explore not only the uncertainty introduced by $\epsilon_{m,k,t}$ but also the uncertainty arising from the estimation of the model $g$.

In practice, updating $g$ at every decision time is unnecessary. Adopting an offline-training-online-deployment paradigm is more suitable, particularly when collecting observations and making decisions in batches. Specifically, we would update the posterior of $g$ at particular time points using all accumulated information. Then, a $\tilde{g}$ is sampled and utilized as the prior for subsequent learning of $\vthe$ until $g$ is retrained and sampled. %For more details on such a computationally efficient variant, refer to Appendix \ref{efficient-variants}.

Given $\tilde{\vthe}_m$, the final step involves an optimization problem. Specifically, we need to solve
$\mathop{\rm{argmax}}\limits_{\boldsymbol{a}_{m,t}\in\mathcal{S}_m} \sum_{k\in [K_m]}\tilde{\theta}_{m,k,a_{m,k,t}},$ which can be regarded as a Multiple-Choice Knapsack Problem (MCKP) \cite{kellerer2004multiple}. Specifically, MCKP is a generalization of the ordinary knapsack problem, where the set of items are originally partitioned into $K$ groups. Instead of making binary choices regarding each item, MCKP only allows (at most) one item in the same group to be chosen. Similar to the ordinary knapsack problem, one can utilize dynamic programming to find the optimal solution. We summarize the entire learning strategy in Algorithm \ref{alg:TS}.

%Since each campaign has independent budget limit $B_m$, we optimize each campaign separately.

\begin{algorithm}[!t]
\SetKwData{Left}{left}\SetKwData{This}{this}\SetKwData{Up}{up}
\SetKwFunction{Union}{Union}\SetKwFunction{FindCompress}{FindCompress}
\SetKwInOut{Input}{Input}\SetKwInOut{Output}{output}
\SetAlgoLined
\Input{Specification of $g$ and the corresponding prior; known parameters (i.e., $\sigma_\epsilon$, $\Cov$); $\mH = \{\}$
}

\For{every decision point j}{ 
Retrieve the campaign index $m$\;

Update the posterior for $g$ as $\prob(g|\mH)$, according to  \eqref{eqn:hierachical_model}\;

Sample a $\tilde{g} \sim \prob(g|\mH)$\;

Given $\tilde{g}$, update the posterior for $\vthe_m$ as $\prob(\vthe_m|\tilde{g}, \mH)$\;

Sample an utility vector $\tilde{\vthe}_{m} \sim \prob(\vthe_m|\tilde{g}, \mH)$\; 

Take action $\boldsymbol{a}_{m,t} = argmax_{\boldsymbol{a}_{m,\cdot}\in\mathcal{S}_m} \sum_{k\in [K_m]}\tilde{\theta}_{m,k,a_{m,k,\cdot}}$\;

Receive reward $R_{m,t}$\;

Update the dataset as $\mH \leftarrow \mH \cup \{(m, \boldsymbol{a}_{m,t}, R_{m,t})\}$
}
\caption{Multi-Task Combinatorial Bandits (MCMAB)}\label{alg:TS}
\end{algorithm}

\section{Offline Evaluation} 

In order to assess the effectiveness of the proposed approach in real-world settings, we compare it with the existing methods using the Amazon Digital Advertisements' campaign data from the first quarter of 2023.

%\textbf{Dataset.} The dataset is sourced from an advertiser that promotes Amazon products and services, which is the largest advertiser and the primary focus of our forthcoming online experiment. To mitigate the influence of extreme values and outliers, we filtered out data points with a cost exceeding 1000 USD. Additionally, we focused on ad lines sourced from the top two suppliers, APS and 3PX, and limited our analysis to six primary channels: Video Mobile App Out-Stream, Video Mobile App In-Stream, Web Billboard, Mobile App, Video Web, and Web. The refined dataset encompasses a total of 39,212 individual observations, which are spread across 1278 ad lines from nine campaigns. By employing the random forest algorithm, we found that the combination of the supply source, channel, and the logarithm of budget cost can effectively account for 94.7\% of the observed variation in the logarithm of clicks obtained (i.e., $log(\theta_{m,k,a})$). This suggests that supply source, channel, and budget cost are crucial factors that significantly impact clicks. However, there are still variations that remain unexplained to us. 
%It should be noted that in order to calculate the random effects, all cost values were converted into integer format.

\textbf{Design.} To simulate real-world scenarios, we first determined $g$, $\sigma_m$, and $\sigma_\epsilon$. The selection of $g$ involved comparing the performance of linear regression, random forest, and CatBoost \citep{prokhorenkova2018catboost} in predicting the logarithm of clicks obtained (i.e., $log(\theta_{m,k,a})$) using campaigns/ad lines' metadata (i.e., supply source, channel) and the logarithm of budget cost. 
CatBoost emerged as the most accurate, exhibiting the lowest mean squared error. Subsequently, let $\Cov = \sigma_m^2\boldsymbol{I}$, $\sigma_m$ and $\sigma_\epsilon$ were determined to be 0.35 and 0.40, respectively, using maximum likelihood estimation based on the fitted CatBoost model and under the assumption that both the random effects and noise are normally distributed. It is important to note that in our application of CatBoost, we do not enforce any parametric modeling assumptions regarding the relationship between the features and the rewards.

To mimic the concurrent scenario, where multiple campaigns run simultaneously, we construct 50 distinct campaigns ($M = 50$), each with five randomly selected ad lines ($K = 5$) and a daily budget limit of \$300 ($B_m = 300$). Budgets were distributed daily across ad lines within each campaign separately. The stochastic observations for the total clicks are generated using the base model (\ref{eqn:hierachical_model}) with ${g,\sigma_m,\sigma_\epsilon}$ as parameters. Similarly, to mimic the sequential scenario, where campaigns come in sequence, each campaign is randomly constructed with five ad lines ($K = 5$) and lasts 50 days ($T = 50$) with a daily budget of \$300 ($B_m = 300$).

\textbf{Baselines.} 
Our studies compare ten approaches, including the proposed MCMAB algorithm with LR, NN, and GP as working models to model the relationship between $log(\theta_{m,k,a})$ and features. We also examine the feature-determined (FD) counterpart of each version of the MCMAB algorithm, which shares the MCMAB's modeling assumptions but with $\sigma_m = 0$. LR-based approaches use feature information $\boldsymbol{x}_{m,k}$, encompassing channels and supply source of the corresponding ad line, and the budget limit of the corresponding campaign. They assume that $g(\boldsymbol{x}_{m,k},a) = \phi(\boldsymbol{x}_{m,k},a)^T\vgamma$, with $\phi$ being a deterministic function that transforms the tuple information $(\boldsymbol{x}_{m,k},a)$ to further include interaction terms between channels, supply sources, and the logarithm of budget shares. GP-based approaches assume $g(\boldsymbol{x}_{m,k},a)$ follows a Gaussian Process, while NN-based approaches employ a fully connected 3-layer neural network $g$ with a width of 30 for the concurrent setting and a width of 26 for the sequential setting.  Additionally, we explored \textit{Hibou}, the current method used in the ADSP system, which posits linear relationships between ad-line performance and assigned budgets, allocating budgets solely on estimated gradients without further exploration or information sharing. The study also includes an evaluation of \textit{Han2021\_RF} and \textit{Han2021\_LR}, the contextual bandits approaches from \citet{han2021budget}, utilizing Random Forest and Linear Regression as the global model, respectively. The local model for each ad line is fitted as a Bayesian linear model using an augmented dataset consisting of 30 predicted returns generated by the fitted global model and the ad line's observed history. Lastly, we considered \textit{FA-ind}, a baseline approach that independently learns the distribution of each $\theta_{m,k,a}$. See Table \ref{baselines} for a summary.
\begin{table}
\centering
\begin{tabular}{ c|c|c|c } 
 \hline
 Models & Utilize $\boldsymbol{x}$ & Heterogeneity & Linear Assumption\\ 
 \hline
 MCMAB (ours)  & $\checkmark$ &  $\checkmark$ & $\times$ \\ 
 FD &  $\checkmark$ &  $\times$& $\times$\\ 
 FA-ind &  $\times$ & $\checkmark$ &  $\times$\\ 
 Han2021 & $\checkmark$ & $\times$ & $\checkmark$\\ 
 \hline
\end{tabular}
\caption{MCMAB and baseline approaches.}\label{baselines}
\vspace{-1.3cm}
\end{table}

To ensure a fair comparison, we applied uninformative priors for all methods. Specifically, for \textit{FD-LR} and \textit{MCMAB-LR}, we used $\vgamma\sim\mathcal{N}(0,20\boldsymbol{I})$; for \textit{FD-GP} and\textit{ MCMAB-GP}, we used zero-mean priors with RBF kernels; for \textit{FD-NN} and \textit{MCMAB-NN}, we initialized the networks with all weights sampled from normal distributions with zero mean \citet{zhang2020neural}; for \textit{Hibou}, we started with an even allocation; and for \textit{Han2021\_LR} and \textit{Han2021\_RF}, we used $\mathcal{N}(0,20\boldsymbol{I})$ as the prior for local model parameters\citep{han2021budget}.

\begin{figure}
    \centering
    \includegraphics[width = \linewidth]{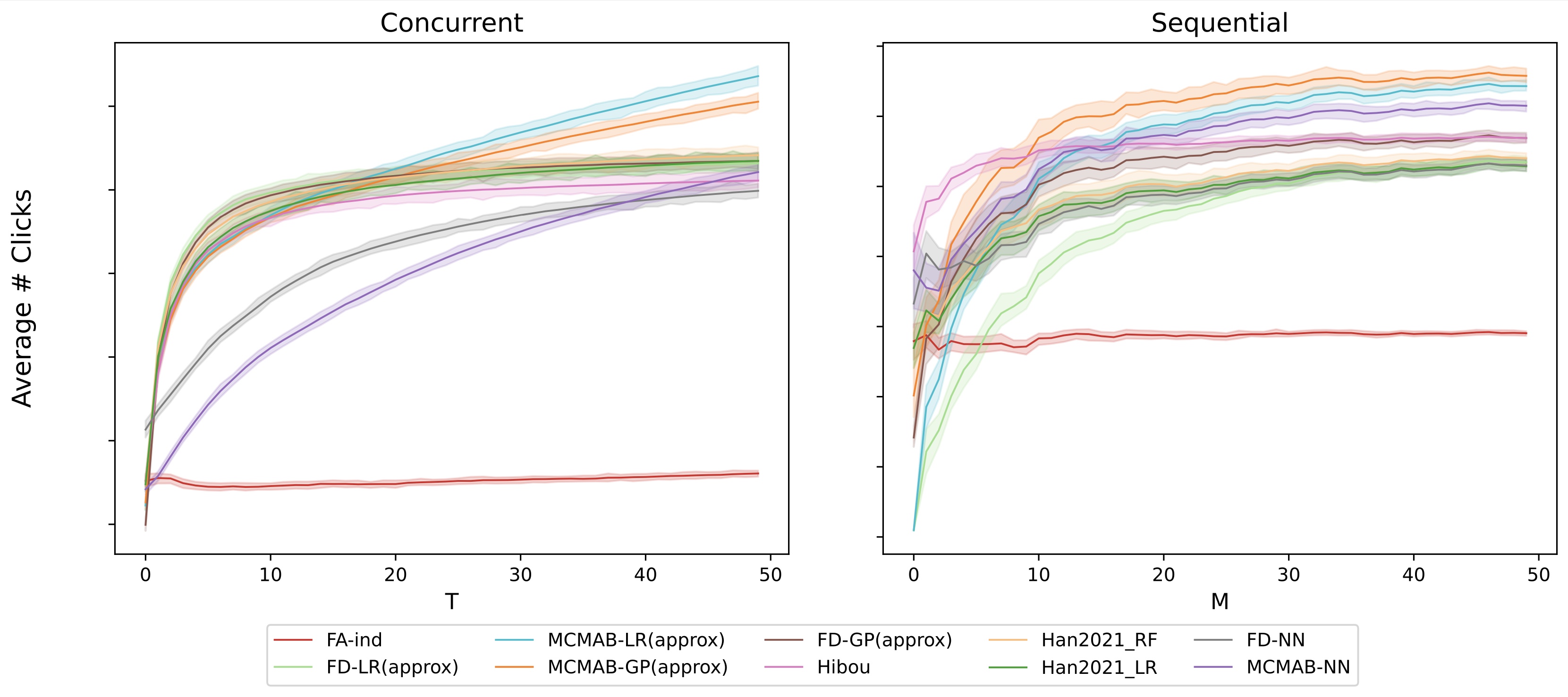}
    \caption{Simulation results on Amazon's campaign data, averaged over 100 random seeds. Shaded areas represent the 95\% CI.}
    \label{simu:real}
    \vspace{-.5cm}
\end{figure}

\textbf{Results.} 
Figure \ref{simu:real} depicts the average reward (i.e., the average number of clicks) received after implementing the budget allocation strategies suggested by each approach. Overall, feature-guided approaches (\textit{MCMAB-LR}, \textit{MCMAB-GP}, and \textit{MCMAB-NN}) demonstrated greater average reward, outperforming other methods. Compared to \textit{Hibou}, \textit{MCMAB} showed an approximate 18\% increase in the average number of clicks obtained at the conclusion of the experiment in the concurrent setting, and a 16\% increase in the sequential setting.

Failing to utilize any feature information, \textit{FA-ind} struggles with the curse of dimensionality and limited interaction opportunities, resulting in a significantly slower learning process with the lowest average reward, for both settings. Under the concurrent setting, \textit{Hibou}, which uses only the budget information and learns the reward distribution for each ad line independently, continues to show a lower average reward than approaches that utilize additional ad line metadata information. On the other hand, feature-determined approaches (\textit{FD-LR}, \textit{FD-GP}) and \textit{Han2021} initially outperform feature-guided approaches (\textit{MCMAB-LR}, \textit{MCMAB-GP}) but ultimately sustain lower average reward due to their restricted model assumptions. It should be noted that because the current network structure is naively specified without carefully fine-tuning its width and depth, \textit{FD-NN} and \textit{MCMAB-NN} perform worse than other feature-determined approaches, indicating that the current network structure fails to capture the relationship's complexity well. We could expect that the NN-based approach will perform better with further fine-tuning.

Under the sequential setting, \textit{Hibou} demonstrates superior performance during the initial stages. This is because when campaigns are introduced sequentially, the metadata information is limited initially, impeding reasonable estimations for other feature-based approaches. As the system accumulates data from an increasingly diverse range of campaigns, the average reward for approaches that utilize metadata for information sharing shows a marked increase. In contrast, \textit{Hibou}'s average reward converges to be constant, reflecting its inability to leverage learnings from past campaigns. Similar to what we observed in the concurrent setting, feature-determined approaches and \textit{Han2021} yield a lower average reward compared to \textit{MCMAB}. This underperformance is primarily attributed to their failure to adequately address the heterogeneity among base-arms.

Finally, \textit{MCMAB-LR} performs better than \textit{MCMAB-NN} and \textit{MCMAB-GP} under the concurrent setting. This is presumably due to the linear properties of the dataset we used, which reduce the efficacy of the more complex GP and NN models, potentially leading to overfitting. In contrast, \textit{MCMAB-GP }performs better than \textit{MCMAB-LR} and \textit{MCMAB-NN} under the sequential setting, with \textit{MCMAB-LR} gradually approaching the performance level of \textit{MCMAB-GP} as more campaigns are completed. This is mainly due to the initial scarcity of metadata, which hinders \textit{MCMAB-LR}'s ability to establish a reliable linear model, whereas the Gaussian process, by using its kernel function, can effectively focus on more relevant features and ignore features containing less information. It is worth noting that fine-tuning kernels and hyperparameters can improve the performance of GP-based approaches, while the performance of NN-based approaches can be enhanced by further adjusting the neural network structure and learning rate.

% using feature > FA
% concurrent: Hibou<FD
% sequential: Hibou>=FD
% FG > Hibou and FD
% concurrent: LR-FG > GP-FG
% sequential: GP-FG >= LR-FG

\vspace{-.2cm}
\section{Online Experiments}
To assess the effectiveness of our method in real-world settings, we initiated an online experiment by allocating 3\% of the budget from an active ad campaign on ADSP. This campaign targeted 15 distinct audience groups, each comprising 22 ad lines. For the experiment, we constructed 30 campaigns, each consisting of a single audience group with 10 ad lines, summing up to 300 ad lines in total. The 3\% of the budget was isolated by carving out 3\% of Amazon demand by geo from the main campaign and directing them exclusively to the experiment.

Within the 3\% of geos in the experiment, we implemented an A/B test, dividing the budget equally between two groups: one subjected to our newly developed MCMAB-based budget allocation strategy and the other continuing with the traditional Hibou budget optimization as a control group. Based on the previous offline evaluation, we selected linear regression as the working model for MCMAB, incorporating features such as the suppliers of the ad lines and the channels used. An informative prior for the MCMAB-LR was constructed using data collected from similar campaigns run by the same advertiser in 2023.

The experiment demonstrated promising potential, achieving a notable 12.7\% reduction in cost-per-click using the proposed method compared to the standard practice after three weeks.
\vspace{-.2cm}
\section{Conclusion}
Motivated by the potential of leveraging information sharing among campaigns for enhancing the accuracy of return predictions, and hence optimizing budget allocation strategies, we introduce a sophisticated multi-task combinatorial bandit framework building upon Bayesian hierarchical models. This innovative approach has demonstrated considerable promise in numerical studies, effectively maximizing cumulative returns through the utilization of the metadata of campaigns and ad lines, as well as the budget size. Specifically, an offline study conducted on Amazon's campaign data reveals an average improvement of 18\% in total clicks obtained under the concurrent setting compared to the currently deployed Hibou system and an average improvement of 16\% under the sequential setting. A further online experiment also shows an improvement of 12.7\% in cost-per-click reduction. %An online evaluation of our proposed system on Amazon campaigns is scheduled for 2024. 
%This will further provide a critical assessment of the framework's efficacy in a real-world context.

As a future direction, recognizing the impact of dynamic contextual factors such as the week of the month, day of the week, and holidays on advertising return variability, we are considering an extension to include contextual bandits. This development aims to incorporate these temporal factors, thereby improving the model's ability to take the seasonality of advertising returns into account. Furthermore, while we assume independence between different campaign activities, internal competition is ubiquitous due to limited ad resources\citep{shen2023cross}. Optimizing each campaign individually may result in a local optimum, with the possibility of excessive demand exceeding ad line supply. To achieve a global optimum, it is worthwhile to consider internal competition across campaigns.

\vspace{-.1cm}
\bibliographystyle{ACM-Reference-Format}
\bibliography{0_main}
\vspace{-.1cm}

%%
%% If your work has an appendix, this is the place to put it.
%\appendix

%\clearpage
%\onecolumn
%\input{6.Appendix1}
%\input{7.Appendix2}
%\input{8.Appendix3}
%\input{9.Appendix4}
\end{document}